\documentclass{colt2017} 
\usepackage{tikz}
\usetikzlibrary{intersections,shapes.arrows}

\title[On the effect of pooling on the geometry of representations]{On the effect of pooling on the geometry of representations}
\usepackage{times}

 \coltauthor{\Name{Gary B{\'e}cigneul} \Email{gary.becigneul@inf.ethz.ch}\\
 \addr Department of Computer Science, ETH Z{\"u}rich and Max Planck ETH Center for Learning Systems
 }

\begin{document}

\maketitle
{\small \noindent\textbf{Editor:} Under Review for COLT 2017\\}

\begin{abstract}
In machine learning and neuroscience, certain computational structures and algorithms are known to yield disentangled representations without us understanding why, the most striking examples being perhaps convolutional neural networks and the ventral stream of the visual cortex in humans and primates. As for the latter, it was conjectured that representations may be disentangled by being flattened progressively and at a local scale \citep{dicarlo2007untangling}. An attempt at a formalization of the role of invariance in learning representations was made recently, being referred to as \textit{I}-theory \citep{anselmi2013unsupervised}. In this framework and using the language of differential geometry, we show that pooling over a group of transformations of the input contracts the metric and reduces its curvature, and provide quantitative bounds, in the aim of moving towards a theoretical understanding on how to disentangle representations.
\end{abstract}


\section{Introduction}
What does disentangling representations mean? In machine learning and neurosciences, representations being tangled has two principal interpretations, and they are intimately connected with each other. The first one is geometrical: consider two sheets of paper of different colors, place one of the two on top of the other, and crumple them together in a paper ball; now, it may look difficult to separate the two sheets with a third one: they are \textit{tangled}, one color sheet representing one class of a classification problem. The second one is analytical: consider a dataset being parametrized by a set of coordinates $\{x_i\}_{i\in I}$, such as images parametrized by pixels, and a classification task between two classes of images. On the one hand, we cannot find a subset $\{x_i\}_{i\in J}$ with $J\subset I$ of this coordinate system such that a variation of these would not change the class of an element, while still spanning a reasonable amount of different images of this class. On the other hand, we are likely to be capable of finding a large amount of transformations preserving the class of any image of the dataset, without being expressible as linear transformations on this coordinate system, and this is another way to interpret representations or factors of variation as being tangled.

Why is disentangling representations important? On the physiological side, the brains of humans and primates alike have been observed to solve object recognition tasks by progressively disentangling their representations \textit{via} the visual stream, from V1 to the IT cortex \citep{dicarlo2007untangling,dicarlo2012does}. On the side of  deep learning, deep convolutional neural networks are also able to disentangle highly tangled representations, since a softmax $-$ which, geometrically, performs essentially a linear separation $-$ computed on the representation of their last hidden layer can yield very good accuracy \citep{krizhevsky2012imagenet}. Conversely, disentangling representations might be sufficient to pre-solve practically any task relevant to the observed data \citep{bengio2013deep}. 

How can we design algorithms in order to move towards more disentangled representations? Although it was conjectured that the visual stream might disentangle representations by flattening them locally, thus inducing a decrease in the curvature globally \citep{dicarlo2007untangling}, the mechanisms underlying such a disentanglement, whether it be for the brain or deep learning architectures, remain very poorly understood \citep{dicarlo2007untangling,bengio2013deep}. However, it is now of common belief that computing representations that are invariant with respect to irrelevant transformations of the input data can help. Indeed, on the one hand, deep convolutional networks have been noticed to naturally learn more invariant features with deeper layers \citep{goodfellow2009measuring, lenc2015understanding,tensmeyer2016improving}. On the other hand, the V1 part of the brain similarly achieves invariance to translations and rotations \textit{via} a ``pinwheels'' structure, which can be seen as a principal fiber bundle \citep{petitot2003neurogeometry,poggio2012computational}. Conversely, enforcing a higher degree of invariance with respect to not only translations, but also rotations, flips, and other groups of transformation has been shown to achieve state-of-the-art results in various machine learning tasks \citep{bruna2013invariant, gens2014deep,oyallon2015deep, dieleman2016exploiting, cohen2016group,cohen2016steerable}, and is believed to help in linearizing small diffeomorphisms \citep{mallat2016understanding}. To the best of our knowledge, the main theoretical efforts in this direction include the theory of scattering operators \citep{mallat2012group,wiatowski2015mathematical} as well as  \textit{I}-theory \citep{anselmi2013unsupervised, anselmi2013magic,anselmi2014representation,anselmi2016invariance}. In particular, \textit{I}-theory permits to use the whole apparatus of kernel theory to build invariant features \citep{mroueh2015learning,raj2016local}.

Our work builds a bridge between the idea that disentangling is a result of \textit{(i)} a local decrease in the curvature of the representations, and \textit{(ii)} building representations that are invariant to nuisance deformations, by proving that pooling over such groups of transformations results in a local decreasing of the curvature. 

We start by providing some background material, after which we introduce our formal framework and theorems, which we then discuss in the case of the non-commutative group generated by translations and rotations.

\section{Some background material}
\subsection{Groups and geometry}
A \textit{group} is a set $G$ together with a map $\cdot:G\times G\rightarrow G$ such that:

\textit{(i)} $\forall g,g',g''\in G,\ g\cdot(g'\cdot g'')=(g\cdot g')\cdot g''$,

\textit{(ii)} $\exists e\in G,\ \forall g\in G,\ g\cdot e=e\cdot g=g$, 

\textit{(iii)} $\forall g\in G,\ \exists g^{-1}\in G:\ g\cdot g^{-1}=g^{-1}\cdot g=e$,\\
where $e$ is called the \textit{identity element}. We write $gg'$ instead of $g\cdot g'$ for simplicity. If, moreover, $gg'=g'g$ for all $g,g'\in G$, then $G$ is said to be \textit{commutative} or \textit{abelian}.\\

A \textit{subgroup} of $G$ is a set $H\subset G$ such that for all $h,h'\in H$, $hh'\in H$ and $h^{-1}\in H$. A subgroup $H$ of a group $G$ is said to be \textit{normal} in $G$ if for all $g\in G$, $gH=Hg$, or equivalently, for all $g\in G$ and $h\in H$, $ghg^{-1}\in H$. If $G$ is abelian, then all of its subgroups are normal in $G$.\\

A \textit{Lie group} is a group which is also a smooth manifold, and such that its product law and inverse map are smooth with respect to its manifold structure. A Lie group is said to be \textit{locally compact} if each of its element possesses a compact neighborhood. On every locally compact Lie group, one can define a \textit{Haar measure}, which is a left-invariant, non-trivial Lebesgue measure on its Borel algebra, and is uniquely defined up to a positive scaling constant. If this Haar measure is also right-invariant, then the group is said to be \textit{unimodular}. This Haar measure is always finite on compact sets, and strictly positive on non-empty open sets. Examples of unimodular Lie groups include in particular all abelian groups, compact groups, semi-simple Lie groups and connected nilpotent Lie groups.\\

A group $G$ is said to be \textit{acting} on a set $X$ if we have a map $\cdot:G\times X\rightarrow X$ such that for all $g,g'\in G$, for all $x\in X$, $g\cdot(g'\cdot x)=(gg')\cdot x$ and $e\cdot x=x$. If this map is also smooth, then we say that $G$ is \textit{smoothly acting} on $X$. We write $gx$ instead of $g\cdot x$ for simplicity. Then, the \textit{group orbit} of $x\in X$ under the action of $G$ is defined by $G\cdot x=\{gx\mid g\in G\}$, and the stabilizer of $x$ by $G_x=\{g\in G\mid gx=x\}$. Note that $G_x$ is always a subgroup of $G$, and that for all $x,y\in X$, we have either $(G\cdot x)\cap(G\cdot y)=\varnothing$, or $G\cdot x=G\cdot y$. Hence, we can write $X$ as the disjoint union of its group orbits, \textit{i.e.} there exists a minimal subset $\tilde{X}\subset X$ such that $X=\sqcup_{x\in\tilde{X}}G\cdot x$. The set of orbits of $X$ under the action of $G$ is written $X/G$, and is in one-to-one correspondence with $\tilde{X}$. Moreover, note that if $H$ is a subgroup of $G$, then $H$ is naturally acting on $G$ via $(h,g)\in H\times G\mapsto hg\in G$; if we further assume that $H$ is normal in $G$, then one can define a canonical group structure on $G/H$, thus turning the canonical projection $g\in G\mapsto H\cdot g$ into a group morphism.\\

A \textit{diffeomorphism} between two manifolds is a map that is smooth, bijective and has a smooth inverse. A \textit{group morphism} between two groups $G$ and $G'$ is a map $\varphi:G\rightarrow G'$ such that for all $g_1,g_2\in G$, $\varphi(g_1g_2)=\varphi(g_1)\varphi(g_2)$. A \textit{group isomorphism} is a bijective group morphism, and a \textit{Lie group isomorphism} is a group isomorphism that is also a diffeomorphism.\\

The \textit{Lie algebra} $\mathfrak{g}$ of a Lie group $G$ is its tangent space at $e$, and is endorsed with a bilinear map $[\cdot,\cdot]:\mathfrak{g}\times\mathfrak{g}\to\mathfrak{g}$ called its \textit{Lie bracket}, and such that for all $x,y,z\in\mathfrak{g}$, $[x,y]=-[y,x]$ and $[x,[y,z]]+[y,[z,x]]+[z,[x,y]]=0$. Moreover, there is a bijection between $\mathfrak{g}$ and left-invariant vector fields on $G$, defined by $\xi\in\mathfrak{g}\mapsto\{g\in G\mapsto d_eL_g(\xi)\}$, where $L_g(h)=gh$ is the left translation. Finally, the flow $t\mapsto\phi_t$ of such a left-invariant vector field $X_{\xi}$ is given by $\phi_t(g)=g\exp(t\xi)$, where $\exp:\mathfrak{g}\to G$ is the exponential map on $G$.\\

For more on Lie groups, Lie algebras, Lie brackets and group representations, see \cite{kirillov2008introduction}, and for a rapid and clear presentation of the notions of sectional curvature and Riemannian curvature, see \cite{andrews2010ricci}.
\subsection{\textit{I}-theory}
\textit{I}-theory aims at understanding how to compute a representation of an image $I$ that is both unique and invariant under some deformations of a group $G$, and how to build such representations in a hierarchical way \citep{poggio2012computational,anselmi2013unsupervised, anselmi2013magic,anselmi2014representation,anselmi2016invariance}.\\

Suppose that we are given a Hilbert space $\mathcal{X}$, typically $L^2(\mathbb{R}^2)$, representing the space of images. Let $G$ be a locally compact group acting on $\mathcal{X}$. Then, note that the group orbit $G\cdot I$ constitutes such an invariant and unique representation of $I$, as $G\cdot I=G\cdot (gI)$, for all $g\in G$, and since two group orbits intersecting each other are equal. \\
 
But how can we compare such group orbits? For an image $I\in\mathcal{X}$, define the map $\Theta_I:g\in G\mapsto gI\in\mathcal{X}$ and the probability distribution $P_I(A)=\mu_G(\Theta_I^{-1}(A))$ for any borel set $A$ of $\mathcal{X}$, where $\mu_G$ is the Haar measure on $G$. For $I,I'\in\mathcal{X}$, write $I\sim I'$ is there exists $g\in G$ such that $I=gI'$. Then, one can prove that $I\sim I'$ if and only if $P_I=P_{I'}$. Hence, we could compare $G\cdot I$ and $G\cdot I'$ by comparing $P_I$ and $P_{I'}$. However, computing $P_I$ can be difficult, so one must be looking for ways to approximate $P_I$. If $t\in\mathbb{S}(L^2(\mathbb{R}^2))$, define $P_{\langle I,t\rangle}$ to be the distribution associated with the random variable $g\mapsto\langle gI,t\rangle$. One can then prove that $P_{I}=P_{I'}$ if and only if $P_{\langle I,t\rangle}=P_{\langle I',t\rangle}$ for all $t\in \mathbb{S}(L^2(\mathbb{R}^2))$, and then provide a lower bound on the sufficient number $K$ of such templates $t_k$, $1\leqslant k\leqslant K$, drawn uniformly on $\mathbb{S}(L^2(\mathbb{R}^2))$, in order to recover the information of $P_I$ up to some error $\varepsilon$ and with high probability $1-\delta$. Finally, each $P_{\langle I,t_k\rangle}$ can be approximated by a histogram $$h^k_n(I)=\dfrac{1}{\vert G\vert}\sum_{g\in G}\eta_n(\langle gI,t_k\rangle),$$ if $G$ is finite or $$h^k_n(I)=\dfrac{1}{\mu_G(G)}\int_{g\in G}\eta_n(\langle gI,t_k\rangle)d\mu_G(g),$$ if $G$ is compact, where $\eta_n$ are various non-negative and possibly non-linear functions, $1\leqslant n\leqslant N$, such as sigmoid, ReLU, modulus, hyperbolic tangent or $x\mapsto\vert x\vert ^p$, among others.\\

In the case where the group $G$ is only partially observable (for instance if $G$ is only locally compact but not bounded), one can define instead a ``partially invariant representation'', replacing each $h^k_n(I)$ by $$\dfrac{1}{\mu_G(G_0)}\int_{g\in G_0}\eta_n(\langle gI,t_k\rangle)d\mu_G(g),$$ where $G_0$ is a compact subset of $G$ which can be observed in practice. Under some ``localization condition'' (see \citep{anselmi2013unsupervised}), it can be proved that this representation is invariant under deformations by elements of $G_0$. When this localization condition is not met, we do not have any exact invariance a priori, but one might expect that the variation in directions defined by the symmetries of $G_0$ is going to be reduced.\\

For instance, let $G$ be the group $\mathbb{R}^2$ of translations in the plane, $G_0=[-a,a]^2$ for some $a>0$, $\eta: x\mapsto (\sigma(x))^2$ where $\sigma$ is a point-wise non-linearity commonly used in neural networks, and $t_k\in\mathbb{S}(L^2(\mathbb{R}^2))$ for $1\leqslant k\leqslant K$. Then, note that the quantities $$\sqrt{\vert G_0\vert h_k(I)}=\sqrt{\sum_{g\in G_0}\eta(\langle gI,t_k\rangle)},$$ for $1\leqslant k\leqslant K$ are actually computed by a 1-layer convolutional neural network with filters $(t_k)_{1\leqslant k\leqslant K}$, non-linearity $\sigma$ and $L^2$-pooling. Moreover, the family $(\sqrt{\vert G_0\vert h_k(gI)})_{g\in G}$ is exactly the output of this convolutional layer, thus describing a direct correspondence between pooling and locally averaging over a group of transformations.\\

Another correspondence can be made between this framework and deep learning architectures. Indeed, assume that during learning, the set of filters of a layer of a convolutional neural network becomes stable under the action of some unknown group $G$ acting on the pixel space, and denote by $\sigma$ the point-wise non-linearity computed by the network. Moreover, suppose that the convolutional layer and point-wise non-linearity are followed by an $L^p$-pooling, defined by $\Pi_\phi^p(I)(x)=\big(\int_{y\in\mathbb{R}^2}\vert I(y)\textbf{1}_{[0,a]^2}(x-y)\vert^p\ dy\big)^{1/p}$. Then, observe that the convolutional layer outputs the following feature maps: $$\{\Pi_\phi^p(\sigma(I\star t_k))\}_{1\leqslant k\leqslant K}.$$ Besides, if the group $G$ has a unitary representation, and if its action preserves $\mathbb{R}^2$, then for all $g\in G$ and $1\leqslant k\leqslant K$, we have $$\Pi_\phi^p(\sigma(gI\star t_k))=\Pi_\phi^p(\sigma(g(I\star g^{-1}t_k)))=g\Pi_\phi^p(\sigma(I\star g^{-1}t_k)).$$ Then, the following layer of the convolutional network is going to compute the sum across channels $k$ of these different signals. However, if our set of filters $t_k$ can be written as $G_0\cdot t$ for some filter $t$ and a subpart $G_0$ of $G$, then this sum will be closely related to a histogram as in \textit{I}-theory:  $$\sum_{g\in G_0}\Pi_\phi^p(\sigma(I\star gt))=\sum_{g\in G_0}g\Pi_\phi^p(\sigma(g^{-1}I\star t_k)).$$
In other words, (local) group invariances are free to appear during learning among filters of a convolutional neural network, and will naturally be pooled over by the next layer. For more on this, see \citep{bruna2013learning,mallat2016understanding}.\\

Finally, let's mention that this implicit pooling over symmetries can also be computed explicitly, and such group invariances across filters enforced, if we know the group in advance, as in G-CNNs and steerable CNNs \citep{cohen2016group,cohen2016steerable}. 

\section{Main results: formal framework and theorems}

Let $G$ be a finite-dimensional, locally compact and unimodular Lie group smoothly acting on $\mathbb{R}^2$. This defines an action $(L_gf)(x)=f(g^{-1}x)$ on $L^2(\mathbb{R}^2)$. Let $G_0$ be a compact neighborhood of the identity element $e$ in $G$, and assume that there exists $\lambda>0$ such that for all $g_0\in G_0$, $\sup_{x\in\mathbb{R}^2}\vert J_{g_0}(x)\vert\leqslant\lambda$, where $J_g$ is the determinant of the Jacobian matrix of $g$ seen as a diffeomorphism of $\mathbb{R}^2$. We define $\Phi:L^2(\mathbb{R}^2)\rightarrow L^2(\mathbb{R}^2)$, the averaging operator on $G_0$, by $$\Phi(f)=\dfrac{1}{\mu_G(G_0)}\int_{g\in G_0}L_gf\ d\mu_G(g).$$ 

 Our first result describes how the euclidean distance in $L^2(\mathbb{R}^2)$ between a function $f$ and its translation by some $g\in G_0$ is contracted by this locally averaging operator.\\

\noindent\textbf{Theorem 1.}\\
For all $f\in L^2(\mathbb{R}^2)$, for all $g\in G$,
$$\Vert\Phi(L_g f)-\Phi(f)\Vert_2\leqslant\sqrt{\lambda}\max\left(1,\sqrt{\Vert J_g\Vert}_{\infty}\right)\dfrac{\mu_G((G_0g)\Delta G_0)}{\mu_G(G_0)}\Vert f\Vert_2.$$\\
\textit{Proof:} See Appendix A.\\

The symbol $\Delta$ above is defined $A\Delta B=(A\cup B)\setminus(A\cap B)=(A\setminus B)\cup(B\setminus A)$. Note that, as one could have expected, this result doesn't depend on the scaling constant of the Haar measure. Intuitively, this result formalizes the idea that locally averaging with respect to some factors of variation, or coordinates, will reduce the variation with respect to those coordinates. The following drawings illustrate the intuition behind Theorem 1, where we pass from left to right by applying $\Phi$.\\\\\\
\begin{minipage}{.2\textwidth}
\resizebox{200pt}{!}{
\begin{tikzpicture}
\path[name path=border1] (0,0) to[out=-10,in=150] (6,-2);
\path[name path=border2] (12,1) to[out=150,in=-10] (5.5,3.2);
\path[name path=redline] (0,-0.4) -- (12,1.5);
\path[name intersections={of=border1 and redline,by={a}}];
\path[name intersections={of=border2 and redline,by={b}}];
\shade[left color=gray!10,right color=gray!80] 
  (0,0) to[out=-10,in=150] (6,-2) -- (12,1) to[out=150,in=-10] (5.5,3.7) -- cycle;
\shade[left color=red!10,right color=red!80] 
  (1.3,0.13) to[out=-10,in=150] (6.3,-1.54) -- (11.3,0.96) to[out=150,in=-10] (5.88,3.21) -- cycle;
\shade[left color=blue!10,right color=blue!80] 
  (0.5,0.1) to[out=-10,in=150] (5.5,-1.57) -- (10.5,0.93) to[out=150,in=-10] (5.08,3.18) -- cycle;
\draw[color=black] (5.5,0.9) to [bend left=20] (6.5,1);
\draw[densely dotted] (5.5,0.9) -- (6.5,1);
\draw[fill=red] (6.5,1) circle (3pt);
\draw[fill=blue] (5.5,0.9) circle (3pt);
\node[rotate=28.6] at (6.2,-1.75) {{\footnotesize $G\cdot f$}};
\node[rotate=28.6] at (5.77,-1.27) {{\footnotesize $G_0\cdot f$}};
\node[rotate=28.6] at (6.95,-1.05) { {\footnotesize $G_0\cdot (L_gf)$}};
\node at (5.5,1.25) {{\footnotesize $f$}};
\node at (6.5,1.35) {{\footnotesize $L_gf$}};
\end{tikzpicture}
}
\end{minipage}
\begin{minipage}{.7\textwidth}
\begin{flushright}
\resizebox{200pt}{!}{
\begin{tikzpicture}
\path[name path=border1] (0,0) to[out=-10,in=150] (6,-2);
\path[name path=border2] (12,1) to[out=150,in=-10] (5.5,3.2);
\path[name path=redline] (0,-0.4) -- (12,1.5);
\path[name intersections={of=border1 and redline,by={a}}];
\path[name intersections={of=border2 and redline,by={b}}];
\shade[left color=gray!10,right color=gray!80] 
  (0,0) to[out=-10,in=150] (6,-2) -- (12,1) to[out=150,in=-10] (5.5,3.7) -- cycle;

\draw[color=black] (5.75,0.9) to [bend left=20] (6.25,1);
\draw[densely dotted] (5.75,0.9) -- (6.25,1);
\draw[fill=red] (6.25,1) circle (3pt);
\draw[fill=blue] (5.75,0.9) circle (3pt);
\node[rotate=28.6] at (6.40,-1.64) {{\footnotesize $\Phi(G\cdot f)$}};
\node at (5.3,1.15) {{\footnotesize $\Phi(f)$}};
\node at (6.3,1.35) {{\footnotesize $\Phi(L_gf)$}};
\end{tikzpicture}
}
\end{flushright}
\end{minipage}\\\\
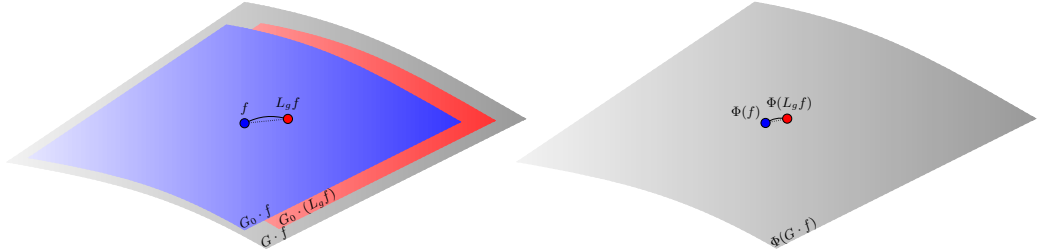
{\scriptsize Figure 1: Concerning the drawing on the left-hand side, the blue and red areas correspond to the compact neighborhood $G_0$ centered in $f$ and $L_gf$ respectively, the grey area represents only a visible subpart of the whole group orbit, the thick, curved line is a geodesic between $f$ and $L_gf$ inside the orbit $G\cdot f$, and the dotted line represents the line segment between $f$ and $L_gf$ in $L^2(\mathbb{R}^2)$, whose size is given by the euclidean distance $\Vert L_gf-f\Vert_2$.}\\

Note that the quantity $\frac{\mu_G((G_0g)\Delta G_0)}{\mu_G(G_0)}$, depending on the geometry of the group, is likely to decrease when we increase the size of $G_0$: if $G=\mathbb{R}^2$ is the translation group, $G_0=[0,a]^2$ for some $a>0$, and $g_\varepsilon$ is the translation by the vector $(\varepsilon,\varepsilon)$, then $\mu_G$ is just the usual Lebesgue measure in $\mathbb{R}^2$ and \[\dfrac{\mu_G((G_0g_\varepsilon)\Delta G_0)}{\mu_G(G_0)}\mathrel{\mathop{\sim}_{\varepsilon\to 0}}2\dfrac{2a\varepsilon}{a^2}=\dfrac{4\varepsilon}{\sqrt{\mu_G(G_0)}}.\]
Indeed, locally averaging over a wider area will decrease the variation even more.\\

As images are handily represented by functions from the space of pixels $\mathbb{R}^2$ to either $\mathbb{R}$ or $\mathbb{C}$, let us define our dataset $\mathcal{X}$ to be a finite-dimensional manifold embedded in a bigger space of functions $\mathcal{Y}$. As for technical reasons we will need our functions to be $L^2$, smooth, and with a gradient having a fast decay at infinity, we choose $\mathcal{Y}$ to be the set of functions $f\in L^2(\mathbb{R}^2)\cap\mathcal{C}^{\infty}(\mathbb{R}^2)$ such that $\vert\langle\nabla f(x),x\rangle\vert=O_{x\to\infty}(\frac{1}{\Vert x\Vert^{1+\varepsilon}})$, for some fixed small $\varepsilon>0$. Note that in practice, images are only non-zero on a compact domain, therefore these assumptions are not restrictive.\\

Further assume that for all $f\in\mathcal{X}$, for all $g\in G$, $L_g f\in\mathcal{X}$. Intuitively, $\mathcal{X}$ is our manifold of images, and $G$ corresponds to the group of transformations that are not relevant to the task at hand. Recall that from \textit{I}-theory, the orbit of an image $f$ under $G$ constitutes a good unique and invariant representation. Here, we are interested in comparing $G\cdot f$ and $\Phi(G\cdot f)$, \textit{i.e.} before and after locally averaging.\\

But how can we compute a bound on the curvature of $\Phi(G\cdot f)$? It is well known that in a Lie group endorsed with a bi-invariant pseudo-Riemannian metric $\langle\cdot,\cdot\rangle$, the Riemann curvature tensor is given by $$R(X,Y,Z,W)=-\dfrac{1}{4}\langle[X,Y],[Z,W]\rangle,$$ where $X,Y,Z,W$ are left-invariant vector-fields, and hence if $(X,Y)$ forms an orthonormal basis of the plane they span, then the sectional curvature is given by $$\kappa(X\wedge Y)=R(X,Y,Y,X)=\dfrac{1}{4}\langle[X,Y],[X,Y]\rangle.$$

Therefore, would we be able to define a Lie group structure and a bi-invariant pseudo-Riemannian metric on $\Phi(G\cdot f)$, we could use this formula to compute its curvature. First, we are going to define a Lie group structure on $G\cdot f$, which we will then transport on $\Phi(G\cdot f)$. As a Lie group structure is made of a smooth manifold structure and a compatible group structure, we need to construct both. In order to obtain the group structure on the orbit, let's assume that the stabilizer $G_f$ is normal; a condition that is met for instance if $G$ is abelian, or if this subgroup is trivial, meaning that $f$ does not have internal symmetries corresponding to those of $G$, which is only a technical condition, as it can be enforced in practice by slightly deforming $f$, by breaking the relevant symmetries with a small noise. Besides, in order to obtain a smooth manifold structure on the orbits, we need to assume that $G_f$ is an embedded Lie subgroup of $G$, which, from proposition B.0 (see appendix), is met automatically when this group admits a finite-dimensional representation.\\

Then, from proposition B.1, there is one and only one manifold structure on the topological quotient space $G/G_f$ turning the canonical projection $\pi:G\rightarrow G/G_f$ into a smooth submersion; moreover, the action of $G$ on $G/G_f$ is smooth, $G/G_f$ is a Lie group, $\pi$ is a Lie group morphism, the Lie algebra $\mathfrak{g}_f$ of $G_f$ is an ideal of the Lie algebra $\mathfrak{g}$ of $G$ and the linear map from $T_e G/T_e G_f$ to $T_{eG_f}(G/G_f)$ induced by $T_e\pi$ is a Lie algebra isomorphism from $\mathfrak{g}/\mathfrak{g}_f$ to the Lie algebra of $G/G_f$.\\

Finally, we need a geometrical assumption on the orbits, insuring that $G$ is warped on $G\cdot f$ in a way that is not ``fractal'', \textit{i.e.} that this orbit can be given a smooth manifold structure: assume that $G\cdot f$ is locally closed in $\mathcal{X}$. Using this assumption and proposition B.2, the canonical map $\Theta_f:G/G_f\rightarrow\mathcal{X}$ defined by $\Theta_f(g G_f)=L_g f$ is a one-to-one immersion, whose image is the orbit $G\cdot f$, which is a submanifold of $\mathcal{X}$; moreover, $\Theta_f$ is a diffeomorphism from $G/G_f$ to $G\cdot f$. Further notice that $\Theta_f$ is $G$-equivariant, \textit{i.e.} for all $g,g'\in G$, $$\Theta_f(g(g'G_f))=L{gg'}f=L_gL_{g'}f=L_g\Theta_f(g'G_f).$$ Moreover, we can define on $G\cdot f$ a group law by $$(L_{g_1} f)\cdot(L_{g_2}f):=L_{g_1g_2}f,$$ for $g_1,g_2\in G$. Indeed, let's prove that this definition doesn't depend on the choice of $g_1,g_2$. Assume that $g_i=a_ib_i$ for $a_i\in G$ and $b_i\in G_f$, $i\in\{1,2\}$. Then, as $G_f$ is normal in $G$, there exists $b_1'\in G_f$ such that $b_1a_2=a_2b_1'$. Then $g_1g_2=a_1a_2b_1'b_2$ and hence $L_{g_1g_2}f=L_{a_1a_2}f$, and this group law is well-defined. Now that $G\cdot f$ is a group, observe that $\Theta_f$ is a group isomorphism from $G/G_f$ to $G\cdot f$. Indeed, it is bijective since it is a diffeomorphism, and it is a group morphism as $$\Theta_f((gG_f)(g'G_f))=\Theta_f((gg')G_f)=L_{gg'}f=(L_gf)\cdot(L_{g'}f)=\Theta_f(gG_f)\cdot\Theta_f(g'G_f).$$ Hence, $G\cdot f$ is also a Lie group, since $G/G_f$ is a Lie group and $\Theta_f:G/G_f\rightarrow G\cdot f$ is a diffeomorphism. Moreover, $Lie(G\cdot f)$ is isomorphic to $\mathfrak{g}/\mathfrak{g}_f$ as a Lie algebra, since they are isomorphic as vector spaces ($\Theta_f$ being an immersion), and by the fact that the pushforward of a diffeomorphism always preserves the Lie bracket.\\
	
	Now that we have defined a Lie group structure on $G\cdot f$, how can we obtain one on $\Phi(G\cdot f)$? Suppose that $\Phi$ is injective on $G\cdot f$ and on $Lie(G\cdot f)$. We can thus define a group law on $\Phi(G\cdot f)$ by:$$\forall g_1,g_2\in G/G_f,\ \Phi(L_{g_1}f)\cdot\Phi(L_{g_2}f):=\Phi(L_{g_1g_2}f).$$ As the inverse function theorem tells us that $\Phi$ is a diffeomorphism from $G\cdot f$ onto its image, $\Phi(G\cdot f)$ is now endorsed with a Lie group structure. However, in order to carry out the relevant calculations, we still need to define left-invariant vector-fields on our Lie group orbits.\\
	
  For all $\xi\in\mathfrak{g}$, define the following left-invariant vector-fields respectively on $G\cdot f$ and $\Phi(G\cdot f)$:
  $$X_{\xi}: L_g f\mapsto\dfrac{d}{dt}_{|t=0}(L_gL_{\exp(t\xi)} f),$$
$$\tilde{X}_{\xi}: \Phi(L_g f)\mapsto\dfrac{d}{dt}_{|t=0}\Phi(L_gL_{\exp(t\xi)}f).$$
We can now state the following theorem:\\\\
\textbf{Theorem 2.}\\
For all $f\in\mathcal{X}$, for all $\xi,\xi'\in\mathfrak{g}$,
$$\Vert[\tilde{X}_\xi,\tilde{X}_{\xi'}]_{\Phi(f)}\Vert_2^2\leqslant\lambda\big[\dfrac{d}{ds}_{|s=0}\dfrac{\mu_G((G_0\exp(s[\xi,\xi']))\Delta G_0)}{\mu_G(G_0)}\big]^2\Vert f\Vert_2^2.$$
\textit{Proof:} See Appendix A.\\

As $\mathcal{X}$ is a manifold embedded in $L^2(\mathbb{R}^2)$, it inherits a Riemannian metric by projection of the usual inner-product of $L^2(\mathbb{R}^2)$ on the tangent bundle of $\mathcal{X}$. Moreover, if we further assume that for all $g\in G$, $\vert J_g\vert=1$, then this Riemannian metric is bi-invariant, and we can finally use the above formula on the Riemannian curvature, together with the previous inequality, to compute a bound on the curvature in a Lie group endorsed with an bi-invariant metric:\\\\
\textbf{Corollary.}\\
For all $f\in\mathcal{X}$, for all $\xi,\xi'\in\mathfrak{g}$,
$$0\leqslant R_{\Phi(f)}(\tilde{X}_\xi,\tilde{X}_{\xi'},\tilde{X}_{\xi'},\tilde{X}_\xi)\leqslant\big[\dfrac{1}{2}\dfrac{d}{ds}_{|s=0}\dfrac{\mu_G((G_0\exp(s[\xi,\xi']))\Delta G_0)}{\mu_G(G_0)}\big]^2\Vert f\Vert_2^2.$$ And if $(\tilde{X}_\xi,\tilde{X}_{\xi'})$ forms an orthonormal basis of the plane they span in $Lie(\Phi(G\cdot f))=\Phi(Lie(G\cdot f))$, then:
$$0\leqslant \kappa_{\Phi(f)}(\tilde{X}_\xi\wedge \tilde{X}_{\xi'})\leqslant\big[\dfrac{1}{2}\dfrac{d}{ds}_{|s=0}\dfrac{\mu_G((G_0\exp(s[\xi,\xi']))\Delta G_0)}{\mu_G(G_0)}\big]^2\Vert f\Vert_2^2.$$\\
\textit{Remark.} The sectional curvature of the basis $(\tilde{X}_\xi,\tilde{X}_{\xi'})$ at $\Phi(f)$ is also the Gaussian curvature of the two-dimensional surface swept out by small geodesics induced by linear combinations of $\tilde{X}_\xi(\Phi(f))$ and $\tilde{X}_{\xi'}(\Phi(f))$.\\

Among well-known finite-dimensional, locally compact and unimodular Lie group smoothly acting on $\mathbb{R}^2$, there are the group $\mathbb{R}^2$ of translations, the compact groups $O(2)$ and $SO(2)$, the euclidean group $E(2)$, as well as transvections, or shears. Moreover, another class of suitable unimodular Lie groups is given by the one-dimensional flows of Hamiltonian systems, which, as deformations of images, could be interpreted as the smooth evolutions of the screen in a video over time, provided that these evolutions can be expressed as group actions on the pixel space.\\

Finally, let's see what Theorem 2 gives us in the case $G=\mathbb{R}^2\ltimes SO(2)$. Note that this group is not commutative, and its curvature form is not identically zero. Let $\theta\in(-\pi,\pi)$, $a>0$, and $G_0=[-\theta,\theta]\times[0,a]^2$. A representation of this group is given by matrices of the form 
\[g(\theta,x,y)= \left( \begin{array}{ccc}
\cos(\theta) & -\sin(\theta) & x \\
\sin(\theta) &  \cos(\theta) & y\\
0            & 0             & 1 \end{array} \right),\]
and a representation of its Lie algebra is given by 
\[ \xi(\zeta,x,y)=\left( \begin{array}{ccc}
0 & -\zeta & x \\
\zeta &  0 & y\\
0            & 0             & 0 \end{array} \right).\]
The Lie bracket is then given by $$[\xi(\zeta,x,y),\xi(\zeta',x',y')]=\xi(\zeta,x,y)\xi(\zeta',x',y')-\xi(\zeta',x',y')\xi(\zeta,x,y)=\xi(0,\zeta'y-\zeta y',\zeta x'-\zeta' x).$$ As the exponential map on the group of translations is the identity map, and as the Haar measure on $\mathbb{R}^2\ltimes SO(2)$ is just the product of the Haar measures on $\mathbb{R}^2$ and $SO(2)$, we have \begin{multline*}
\mu_G((G_0\exp(s[\xi(\zeta,x,y),\xi(\zeta',x',y']))\Delta G_0)=\\
2\theta\mu_{\mathbb{R}^2}(([s(\zeta'y-\zeta y'),s(\zeta'y-\zeta y')+a]\times[s(\zeta x'-\zeta' x),s(\zeta x'-\zeta' x)+a])\Delta [0,a]^2),
\end{multline*}
and $\mu_G(G_0)=2\theta a^2$. Therefore, when $s\to 0$, we have 
\begin{multline*}
\mu_G((G_0\exp(s[\xi(\zeta,x,y),\xi(\zeta',x',y']))\Delta G_0)
\sim\\ 2\theta\times2(as(\zeta'y-\zeta y')+as(\zeta x'-\zeta' x))=4\theta as(\zeta(x'-y')-\zeta'(x-y)),
\end{multline*} 
from what we deduce that $$\big[\dfrac{1}{2}\dfrac{d}{ds}_{|s=0}\dfrac{\mu_G((G_0\exp(s[\xi(\zeta,x,y),\xi(\zeta',x',y']))\Delta G_0)}{\mu_G(G_0)}\big]^2=\dfrac{(\zeta(x'-y')-\zeta'(x-y))^2}{a^2}.$$
As a consequence, if $f\in\mathcal{X}$ is an image in our dataset, of $L^2$-norm equal to $1$, and if we choose $\xi(\zeta,x,y)$ and $\xi(\zeta',x',y')$ such that the $L^2$ functions $\tilde{X}_{\xi(\zeta,x,y)}(\Phi(f))$ and $\tilde{X}_{\xi(\zeta',x',y')}(\Phi(f))$ are orthogonal in $L^2$ and have $L^2$-norm equal to $1$, then the Gaussian curvature $\kappa$ of the 2-dimensional surface swept out by these two vector fields around $\Phi(f)$, in the Lie group $\Phi(G\cdot f)$, is smaller than: $$\kappa\leqslant\dfrac{(\zeta(x'-y')-\zeta'(x-y))^2}{a^2}.$$

\section{Conclusion}
Being able to disentangle highly tangled representations is a very important and challenging problem in machine learning. In deep learning in particular, there exist successful algorithms that may disentangle highly tangled representations in some situtations, without us understanding why. Similarly, the ventral stream of the visual cortex in humans and primates seems to perform such a disentanglement of representations, but, again, the reasons behind this process are difficult to understand. It is believed that making representations invariant to some nuisance deformations, as well as locally flattening them, might help or even be an essential part of the disentangling process. As shown by our theorems, there is a connection between these two intuitions, in the sense that achieving a higher degree of invariance with respect to some group transformations will flatten the representations in directions of the tangent space corresponding to the Lie algebra generators of these transformations. Using our theorems, we showed that in the case of the group of positive affine isometries, a precise bound on the sectional curvature can be computed, with respect to the pooling parameters. We hope that this work will encourage the geometrical study of how representations evolve during learning, in function of the hyperparameters of the algorithm that is used on these representations.

\acks{We thank Simon Janin and Victor Godet for interesting conversations.}

\nocite{*}
\bibliography{bibliography}

\appendix

\section{Proofs of Theorem 1, Theorem 2 and Lemma}
\noindent\textbf{Theorem 1.}\\
For all $f\in L^2(\mathbb{R}^2)$, for all $g\in G$,
$$\Vert\Phi(L_g f)-\Phi(f)\Vert_2\leqslant\sqrt{\lambda}\max(1,\sqrt{\Vert J_g\Vert}_{\infty})\dfrac{\mu_G((G_0g)\Delta G_0)}{\mu_G(G_0)}\Vert f\Vert_2.$$\\
\textit{Proof:}\\\\
We have $$\mu_G(G_0)^2\Vert \Phi(L_g f)-\Phi(f)\Vert_2^2=\int_{x\in\mathbb{R}^2}\big(\int_{g'\in G_0}L_{g'g} f(x)-L_{g'}f(x)d\mu_G(g')\big)^2 dx,$$ but $$\int_{g'\in G_0}(L_{g'g}f(x)-L_{g'}f(x))d\mu_G(g')=\int_{g'\in G_0}L_{g'g}f(x)d\mu_G(g')-\int_{g'\in G_0}L_{g'}f(x)d\mu_G(g'),$$ \textit{i.e.} setting $g''=g'g$ and using the right-invariance of $\mu_G$, $$\int_{g'\in G_0}(L_{g'g}f(x)-L_{g'}f(x))d\mu_G(g')=\int_{g''\in G_0g}L_{g''}f(x)d\mu_G(g'')-\int_{g'\in G_0}L_{g'}f(x)d\mu_G(g').$$ And using $\int_A h-\int_B h=(\int_{A\setminus B}h+\int_{A\cap B}h)-(\int_{B\setminus A} h+\int_{B\cap A} h)=\int_{A\setminus B}h-\int_{B\setminus A} h$, we have $$\int_{g'\in G_0}(L_{g'g}f(x)-L_{g'}f(x))d\mu_G(g')=\int_{g'\in G_0g\setminus G_0}L_{g'}f(x)d\mu_G(g')-\int_{g'\in G_0\setminus G_0g}L_{g'}f(x)d\mu_G(g').$$ Plugging this in the first equation gives $$\mu_G(G_0)\Vert\Phi(L_gf)-\Phi(f)\Vert_2=\Vert\int_{g'\in G_0g\setminus G_0}(L_{g'}f)d\mu_G(g')-\int_{g'\in G_0\setminus G_0g}(L_{g'}f)d\mu_G(g')\Vert_2,$$ \textit{i.e.} using a triangle inequality $$\mu_G(G_0)\Vert\Phi(L_gf)-\Phi(f)\Vert_2\leqslant\Vert\int_{g'\in G_0g\setminus G_0}(L_{g'}f)d\mu_G(g')\Vert_2+\Vert\int_{g'\in G_0\setminus G_0g}(L_{g'}f)d\mu_G(g')\Vert_2.$$ Now observe that by interverting the integrals using Fubini's theorem, 
$$\Vert\int_{g'\in G_0\setminus G_0g}(L_{g'}f)d\mu_G(g')\Vert_2=\sqrt{\int_{g_1\in G_0\setminus G_0g}\int_{g_2\in G_0\setminus G_0g}\big(\int_{x\in\mathbb{R}^2}(L_{g_1}f)(x)(L_{g_2}f)(x)dx\big) d\mu_G(g_1)d\mu_G(g_2)},$$ and using a Cauchy-Schwarz inequality, 
$$\Vert\int_{g'\in G_0\setminus G_0g}(L_{g'}f)d\mu_G(g')\Vert_2\leqslant\sqrt{\int_{g_1\in G_0\setminus G_0g}\int_{g_2\in G_0\setminus G_0g}\Vert L_{g_1}f\Vert_2\Vert L_{g_2}f\Vert_2 d\mu_G(g_1)d\mu_G(g_2)}.$$ As for all $g'\in G_0$ we have $\Vert L_{g'} f\Vert_2=\Vert f\sqrt{\vert J_{g'}\vert}\Vert_2\leqslant\sqrt{\lambda}\Vert f\Vert_2$ with a change of variables, we have $$\Vert\int_{g'\in G_0\setminus G_0g}(L_{g'}f)d\mu_G(g')\Vert_2\leqslant\sqrt{\lambda}\mu_G(G_0\setminus (G_0g))\Vert f\Vert_2.$$ For the other term, note that by setting $g''=g'g^{-1}$,  we have
$$\Vert\int_{g'\in G_0g\setminus G_0}(L_{g'}f)d\mu_G(g')\Vert_2=\Vert\int_{g''\in G_0\setminus G_0g^{-1}}(L_{g''g}f)d\mu_G(g'')\Vert_2=\Vert\int_{g''\in G_0\setminus G_0g^{-1}}(L_{g''}L_{g}f)d\mu_G(g'')\Vert_2,$$ and then similarly,
$$\Vert\int_{g''\in G_0\setminus G_0g^{-1}}(L_{g''}L_{g}f)d\mu_G(g'')\Vert_2=\sqrt{\int_{g_1\in G_0\setminus G_0g^{-1}}\int_{g_2\in G_0\setminus G_0g^{-1}}\Vert L_{g_1}L_gf\Vert_2\Vert L_{g_2}L_gf\Vert_2 d\mu_G(g_1)d\mu_G(g_2)}.$$ As for all $g'\in G_0$ we have $\Vert L_{g'}L_g f\Vert_2=\Vert f\sqrt{\vert J_{g'g}\vert}\Vert_2\leqslant\sqrt{\lambda\Vert J_g\Vert}_\infty\Vert f\Vert_2$, we have
$$\Vert\int_{g'\in G_0g\setminus G_0}(L_{g'}f)d\mu_G(g')\Vert_2\leqslant\sqrt{\lambda\Vert J_g\Vert}_\infty\mu_G(G_0\setminus (G_0g^{-1}))\Vert f\Vert_2.$$ Therefore
$$\mu_G(G_0)\Vert\Phi(L_gf)-\Phi(f)\Vert_2\leqslant\sqrt{\lambda\Vert J_g\Vert}_\infty\mu_G(G_0\setminus (G_0g^{-1}))\Vert f\Vert_2+\sqrt{\lambda}\mu_G(G_0\setminus (G_0g))\Vert f\Vert_2,$$ and the following fact concludes the proof: $$\mu_G(G_0\setminus (G_0g^{-1}))+\mu_G(G_0\setminus (G_0g))=\mu_G((G_0g)\setminus G_0)+\mu_G(G_0\setminus (G_0g))=\mu_G((G_0g)\Delta G_0).$$
\begin{flushright}
$\square$
\end{flushright}
\textbf{Theorem 2.}\\
For all $f\in\mathcal{X}$, for all $\xi,\xi'\in\mathfrak{g}$,
$$\Vert[\tilde{X}_\xi,\tilde{X}_{\xi'}]_{\Phi(f)}\Vert_2^2\leqslant\lambda\big[\dfrac{d}{ds}_{|s=0}\dfrac{\mu_G((G_0\exp(s[\xi,\xi']))\Delta G_0)}{\mu_G(G_0)}\big]^2\Vert f\Vert_2^2.$$
\textit{Proof:}\\\\
As $\Phi$ realizes a diffeomorphism from $G\cdot f$ onto its image, and as $\Phi$ equals its differential from \textit{Lemma}, we have that for all vector field $X$ on $G\cdot f$, $\Phi_*(X)(\Phi(f))=(d\Phi)_f(X(f))=\Phi(X(f))$. Hence
\begin{align*}
[\tilde{X}_\xi,\tilde{X}_{\xi'}]_{\Phi(f)}&=&[\Phi(X_\xi),\Phi(X_{\xi'})]_f\\
&=&[\Phi(X_\xi)\circ\Phi^{-1},\Phi(X_{\xi'})\circ\Phi^{-1}]_{\Phi(f)}\\
&=&[\Phi_*(X_\xi),\Phi_*(X_{\xi'})]_{\Phi(f)}\\
&=&\Phi_*([X_\xi,X_{\xi'}])(\Phi(f))\\
&=&\Phi([X_\xi,X_{\xi'}]_f).
\end{align*}
Recall that the Lie bracket of left-invariant vector fields is given by the opposite of the Lie bracket of their corresponding generators, hence in our case: $$[X_\xi,X_{\xi'}]=X_{-[\xi,\xi']}=-X_{[\xi,\xi']}.$$
Therefore, 
\begin{align*}
\Vert [\tilde{X}_\xi,\tilde{X}_{\xi'}]_{\Phi(f)}\Vert_2&=&\Vert\Phi([X_\xi,X_{\xi'}]_f)\Vert_2\\
&=&\Vert\Phi(X_{[\xi,\xi']}(f))\Vert_2\\
&=&\Vert\Phi(\lim_{t\to 0}\dfrac{1}{t}(L_{\exp(t[\xi,\xi'])}f-f))\Vert_2\\
&=&\Vert\lim_{t\to 0}\dfrac{1}{t}\big(\Phi(L_{\exp(t[\xi,\xi'])}f)-\Phi(f)\big)\Vert_2.
\end{align*}
From \textit{Theorem 1}, we have 
$$\Vert\Phi(L_{\exp(t[\xi,\xi'])}f)-\Phi(f)\Vert_2\leqslant\sqrt{\lambda}\max(1,\sqrt{\Vert J_{\exp(t[\xi,\xi'])}\Vert}_{\infty})\dfrac{\mu_G((G_0\exp(t[\xi,\xi']))\Delta G_0)}{\mu_G(G_0)}\Vert f\Vert_2.$$ 
As $\exp(t[\xi,\xi'])\to e$ when $t\to 0$, its Jacobian goes to $1$. Moreover, as $f$ has a gradient with fast decay, we can take the limit out of the $L^2$-norm, which concludes the proof.
\begin{flushright}
$\square$
\end{flushright}
\textbf{Lemma.}\\
For all $f\in\mathcal{X}$ and $\xi\in\mathfrak{g}$, $$\dfrac{d}{dt}_{|t=0}\Phi(L_{\exp(t\xi)}f)=\Phi\big(\dfrac{d}{dt}_{|t=0}(L_{\exp(t\xi)}f)\big).$$\\
\textit{Proof:}\\\\
For all $x\in\mathbb{R}^2$, 
\begin{align*}
\big(\dfrac{d}{dt}_{|t=0}\Phi(L_{\exp(t\xi)}f)\big)(x)&=&\dfrac{d}{dt}\big(\dfrac{1}{\mu_G(G_0)}\int_{g'\in G_0}(L_{g'}L_{\exp(t\xi)}f)(x)d\mu_{G_0}(g')\big)_{|t=0}\\
&=&\dfrac{1}{\mu_G(G_0)}\int_{g'\in G_0}\dfrac{d}{dt}\big(f(\exp(-t\xi)g'^{-1}x)\big)_{|t=0}d\mu_{G_0}(g')\\
&=&\dfrac{1}{\mu_G(G_0)}\int_{g'\in G_0}d_{(g'^{-1}x)}f\big(-\xi(g'^{-1}x)\big)d\mu_{G_0}(g')\\
&=&\Phi\big(d_{\cdot}f\big(-\xi(\cdot)\big)\big)(x)\\
&=&\Phi\big(\dfrac{d}{dt}_{|t=0}(L_{\exp(t\xi)}f)\big)(x).
 \end{align*}
 \begin{flushright}
 $\square$
 \end{flushright}
\section{Supplementary material}

The next three propositions are taken from the publicly available french textbook \cite{paulin2014groupes}, in which they're respectively numbered as E.7, 1.60, 1.62.\\\\
\textbf{Proposition B.0} Let $G$ be a Lie group and $\rho:G\rightarrow GL(V)$ a finite-dimensional Lie group representation of $G$. Then for all $v\in V$, the map defined by $g\in G\mapsto\rho(g)v$ has constant rank, and the stabilizer $G_v$ is an embedded Lie subgroup of $G$.\\\\
\textbf{Proposition B.1}. Let $G$ be a Lie group, $H$ be an embedded Lie subgroup of $G$, and $\pi:G\rightarrow G/H$ be the canonical projection. There exists one and only one smooth manifold structure on the topological quotient space $G/H$ turning $\pi$ into a smooth submersion. Moreover, the action of $G$ on $G/H$ is smooth, and if $H$ is normal in $G$, then $G/H$ is a Lie group, $\pi$ is a Lie group morphism, the Lie algebra $\mathfrak{h}$ of $H$ is an ideal of the Lie algebra $\mathfrak{g}$ of $G$ and the linear map from $T_e G/T_e H$ to $T_{eH}(G/H)$ induced by $T_e\pi$ is a Lie algebra isomorphism from $\mathfrak{g}/\mathfrak{h}$ to the Lie algebra of $G/H$.\\\\
\textbf{Proposition B.2}. Let $M$ be a manifold together with a smooth action of a Lie group $G$, and $x\in M$; \textit{(i)} the canonical map $\Theta_x:G/G_x\rightarrow M$ defined by $\Theta_x(gG_x)=gx$ is a one-to-one immersion, whose image is the orbit $G\cdot x$; \textit{(ii)} the orbit $G\cdot x$ is a submanifold of $M$ if and only if it is locally closed in $M$; \textit{(iii)} if $G\cdot x$ is locally closed, then $\Theta_x$ is a diffeomorphism from $G/G_x$ to $G\cdot x$.\\\\

\end{document}